\documentclass{ws-jmrr}
\usepackage{caption}
\usepackage{cite}
\usepackage{amsmath,amssymb,amsfonts}
\usepackage{algorithmic}
\usepackage{graphicx,adjustbox}
\usepackage{textcomp}
\usepackage{xcolor}
\usepackage{color}
\usepackage{afterpage}
\usepackage{makecell} 

\usepackage{colortbl}
\usepackage{float}
\usepackage[font=small,labelfont=bf]{caption}
\usepackage{wrapfig,lipsum,booktabs}
\usepackage{subfig}
\usepackage{units}

\def\BibTeX{{\rm B\kern-.05em{\sc i\kern-.025em b}\kern-.08em
    T\kern-.1667em\lower.7ex\hbox{E}\kern-.125emX}}
\usepackage{multirow}
\usepackage{xspace}

\newcommand*{\ie}{\emph{i.e.}\@\xspace}

\usepackage[acronym]{glossaries}
\newacronym{cao}{CAO}{central airway obstruction}
\newacronym{sam}{SAM}{Segment Anything Model}

\usepackage{fancyhdr}
\renewcommand{\sectionmark}[1]{} 
\fancypagestyle{plain}{ 
    \fancyhf{}
    \fancyhead[R]{\thepage}
    
}

\fancypagestyle{firstpage}{
    \fancyhf{} 
}

\AtBeginDocument{
    
    \footnotetext{Preprint of an article accepted for publication in the Journal of Medical Robotics Research, 2025. Copyright World Scientific Publishing Company [https://worldscientific.com/worldscinet/jmrr].}
}

\begin{document}

\thispagestyle{firstpage} 

\definecolor{antiquefuchsia}{rgb}{0.57, 0.36, 0.51}
\definecolor{teal}{rgb}{0.196, 0.659, 0.620}
\definecolor{TEE_blue}{RGB}{0,128,255}
\definecolor{nicepurple}{RGB}{150,50,200}
\newcommand{\MS}[1]{\textcolor{nicepurple}{[MS: #1]}}
\newcommand{\NY}[1]{\textcolor{TEE_blue}{[NY: #1]}}
\newcommand{\TW}[1]{\textcolor{teal}{[TW: #1]}}
\newcommand{\PS}[1]{\textcolor{green}{[PS: #1]}}
\newcommand{\JG}[1]{\textcolor{red}{[JG: #1]}}
\newcommand{\AKZ}[1]{\textcolor{olive}{[Axel: #1]}} 
\newcommand{\AK}[1]{\textcolor{cyan}{[Alan: #1]}}

\title{Autonomous Vision-Guided Resection of Central Airway Obstruction}

\author{Mariana E. Smith$^{a*}$, Nural Yilmaz$^{a*}$, Tanner Watts$^{b}$, Paul Maria Scheikl$^{a}$, Jiawei Ge$^{a}$, Anton Deguet$^{c}$, Alan Kuntz$^{b}$, and Axel Krieger$^{a}$ }

\address{$^{*}$ Both authors contributed equally to this work.}

\address{$^a$Department of Mechanical Engineering, Johns Hopkins University, Baltimore, MD 21211, USA\\
E-mail: msmit458@jhu.edu}

\address{$^b$Robotics Center and Kahlert School of Computing, University of Utah, Salt Lake City, UT 84112, USA}

\address{$^c$Malone Center for Engineering and Healthcare, Johns Hopkins University, Baltimore, MD 21211, USA}


\maketitle

\begin{abstract}

Existing tracheal tumor resection methods often lack the precision required for effective airway clearance, and robotic advancements offer new potential for autonomous resection. We present a vision-guided, autonomous approach for palliative resection of tracheal tumors. This system models the tracheal surface with a fifth-degree polynomial to plan tool trajectories, while a custom Faster R-CNN segmentation pipeline identifies the trachea and tumor boundaries. The electrocautery tool angle is optimized using handheld surgical demonstrations, and trajectories are planned to maintain a 1\,mm safety clearance from the tracheal surface. We validated the workflow successfully in five consecutive experiments on ex-vivo animal tissue models, successfully clearing the airway obstruction without trachea perforation in all cases (with more than 90\% volumetric tumor removal). These results support the feasibility of an autonomous resection platform, paving the way for future developments in minimally-invasive autonomous resection.

\end{abstract}

\keywords{tumor resection, autonomous surgery, vision-based control}

\begin{multicols}{2}
\section{Introduction}

\Gls{cao} refers to an occlusion of the trachea most commonly caused by tumor extension into the airway~\cite{Chen2011}. In the United States, malignant growths cause \gls{cao} in 80,000 new patients every year, posing significant morbidity~\cite{Chen1998}. Of all patients diagnosed with \gls{cao}, 20\% will suffer from cough, shortness of breath, and obstructive pneumonia~\cite{Ernst2004}. In acute cases, patients are unable to breathe due to full airway occlusion, requiring emergency resection and stabilization. Most patients diagnosed with \gls{cao} also present with advanced lung cancer. In this context, the primary purpose of addressing \gls{cao} is to provide palliative care aimed at enhancing quality of life by improving respiratory function~\cite{Morris2002}.

To avoid the risks associated with open surgery, surgeons address \gls{cao} by deploying a rigid hand-held bronchoscope through the patient’s mouth. A widely employed surgical technique is the `core-out' technique, wherein the surgeon uses the bevel of the bronchoscope to scrape the \gls{cao} from the tracheal wall~\cite{Mathisen1989, Vishwanath2013}. After the tumor is re-

\begin{figurehere}
    \centering
    \includegraphics[width=0.85\columnwidth]{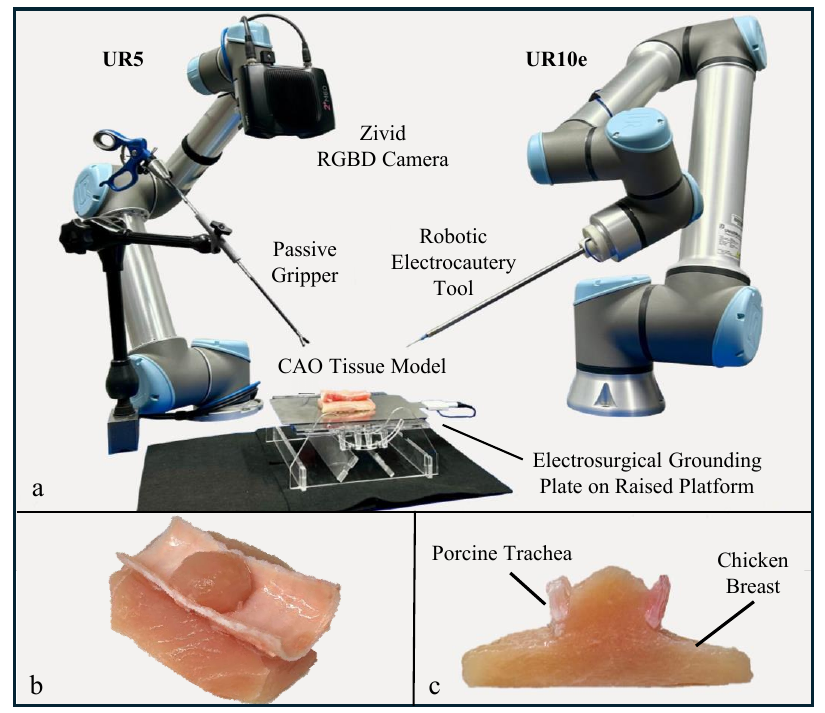}
    \caption{Full system for resection comprised of Zivid camera, passive gripper, and robotic electrocautery tool (a), ex-vivo CAO tissue model (b), and cross-sectional view of tissue model (c).}
    \label{fig:intro}
\end{figurehere}

\noindent  moved, instruments and stents can be deployed through the inner lumen of the bronchoscope~\cite{Ernst2004}. However, the core-out technique requires frequent repositioning and tilting of the bronchoscope, using the patient’s mouth as the fulcrum. This maneuver places significant stress on the patient’s mouth and neck, leading to complications such as spinal injury in approximately one-third of cases~\cite{Vishwanath2013}.

Precise electrocautery is an alternative to the core-out technique, to minimize trauma to the trachea during \gls{cao} removal. This approach was successfully demonstrated by Gafford et al., who performed minimally-invasive \gls{cao} removal with a teleoperated robotic bronchoscope equipped with a monocular endoscope and two concentric tube manipulators for gripping and electrocautery~\cite{Gafford2020}. The \gls{cao} removal was performed on cadaver tissue, showing a significant reduction in applied force to the patient compared to the core-out technique.

However, the robotic system must still be operated by a human surgeon. To address the current shortage of surgeons, recent studies have introduced autonomy in robotic surgical procedures to reduce surgeon workload, particularly during time-consuming or repetitive tasks. Several studies have focused on autonomous robotic tumor resection in the past decade, including removal of foam tumor fragments~\cite{Kehoe2014}, suction of gelatinous fluids~\cite{Hu2018}, resection of a rubber tumor model~\cite{McKinley2016}, and electrosurgery of pseudo-tumors in porcine tissue~\cite{Opfermann2017}. Most recently, a vision-guided robotic system demonstrated the first supervised autonomous tongue tumor resection (\ie, glossectomy) using animal tissues~\cite{Ge2019, Ge2021, Ge2024}. Reported accuracy was on par with manual resections performed by an expert surgeon. However, no studies have yet shown an autonomous resection of \gls{cao}. This work is inspired by the demonstrated success of autonomous glossectomy and by the potential of minimally-invasive robots to perform precise resection within the trachea. This study represents the first demonstration of a supervised autonomous vision-guided workflow for the surgical resection of \gls{cao}.

Development of an autonomous system for \gls{cao} removal requires addressing challenges in perception, planning, robot control, and miniaturization. In this paper, we address the challenges of perception and planning by using a simplified setup. We use rigid-link robotic arms and an open-surgery environment to validate our vision-based autonomous workflow. Our workflow enables straightforward transfer to systems that address miniaturization and control challenges. In future work, the workflow in this paper may be adapted to minimally-invasive manipulators capable of operating within an enclosed trachea.

The contributions of this work are
1) development of an ex-vivo \gls{cao} animal tissue model designed for an open-surgery approach, which allows the use of large rigid-link robots while maintaining realistic tissue effects, and
2) open-surgery demonstration of the first vision-based supervised autonomous workflow for the resection of \gls{cao} in five consecutive models.

\section{Methods}

The full system (Fig. \ref{fig:intro}a) is comprised of a robot-mounted electrocautery tool, a robot-mounted RGBD camera, and a passive laparoscopic gripper. A tissue model is placed on an electrosurgical grounding platform between the robots. The passive gripper tensions the tumor while the robot-mounted electrocautery tool performs the autonomous cut with visual guidance from the mounted RGBD camera, enabling supervised fully autonomous resection of \gls{cao}.

\subsection{Robotic Hardware}

Two UR manipulators (Universal Robots, Odense, Denmark) are used in this study. The UR10e manipulator is equipped with an electrosurgical instrument, which was custom-built from a monopolar electrode (Bovie, Clearwater, FL) with a laparoscopic extension. The UR5 robot holds the RGBD camera (Zivid 2 Plus M60, Oslo, Norway) for imaging of the surgical scene. The Zivid 2 Plus M60 RGBD camera employs structured light 3D imaging and is factory-calibrated. A standard hand-eye calibration procedure was performed with the Zivid camera to obtain the necessary transformation for system registration. A fiducial-based calibration was conducted to obtain the transformation between the UR5 and the UR10e. As in the minimally-invasive case, the electrocautery tool in this work performs each cut while approaching from the distal end of the trachea, as if inserted through the mouth. However, due to the size of the rigid-link robots, the gripper and camera cannot also approach from the distal end; instead, we place the camera above the tumor and the gripper approaching from the proximal end. However, the vision-based resection workflow is not reliant on this configuration and would still be viable if it was transferred to smaller manipulators which all approach from the distal end. Also, since this study focuses on the automation of the resection task alone, the gripper holding the tumor in the other arm was not automated. For the experiments in this paper, a laparoscopic gripper (Snowden-Pencer SP90-6379, Tucker, GA) was fixed on a passive arm and manually translated incrementally along its primary axis to move the tumor before each cut took place. Automating this aspect is left to future work.

\subsection{Surgical Process Modeling}
Before designing the robotic control workflow, it is necessary to define the surgical subtasks which make up the complete procedure of \gls{cao} resection. We adopt a subtask-level Surgical Process Model (SPM) in our approach, inspired by \cite{Nagy2018}. Using SPM representations, the procedure can be split into three subtasks: \textit{Reach-in}, \textit{Resect}, and \textit{Retract}. A finite state machine (FSM) approach naturally follows from these discrete subtasks. FSMs are widely used in autonomous processes due to their maintainable, intuitive structure \cite{FSMs}. In this work, the \textit{Reach-in}, \textit{Resect}, and \textit{Retract} subtasks become finite states, where specific events are programmed to trigger state transitions.

\subsection{Control Workflow}

In our autonomous resection framework (Fig. \ref{fig:workflow}), we aim to provide a systematic approach to supervised tumor resection with point cloud data and polynomial surface fitting. First, data acquisition is conducted as the system uses the Zivid RGBD camera to obtain a snapshot of the surgical scene (which contains a point cloud, color image, depth image, and camera intrinsics). This snapshot data is fed to the segmentation node, which autonomously generates bounding boxes around the tumor and trachea in the color image, and segmentations of the tumor and trachea in the point cloud. These bounding boxes and segmentation are plotted for visualization by the supervisor. The program then prompts the supervisor to either accept the autonomous segmentation or to reject it and hand-draw the bounding boxes, which are then used for autonomous segmentation. The segmented trachea and tumor are published as point clouds to independent ROS topics. 

Next, the trachea surface is fitted with a polynomial surface model. This fitted surface provides a foundation for trajectory planning, where the cutting paths are generated along said surface. After a cut path has been generated, it is displayed to the human supervisor in RViz, along with the RMSE between this cut and the cut which was predicted using the dataset (surface fit and point clouds) stored from the previous cut. The RMSE between cuts indicates how much the newly measured tumor shape and cut trajectory differs from the prior model's predictions (see Section 2.7 for a more detailed explanation of RMSE calculations). If the RMSE value exceeds a set threshold, the system self-supervises by displaying an error to the human operator; the human operator can then choose to re-image and re-plan the cut if desired.

This process is divided into three phases of execution using a finite-state approach: \textit{Reach-in}, \textit{Resect}, and \textit{Retract}. 

\vspace{1em}
\begin{figurehere}
    \centering
    \includegraphics[width=\columnwidth]{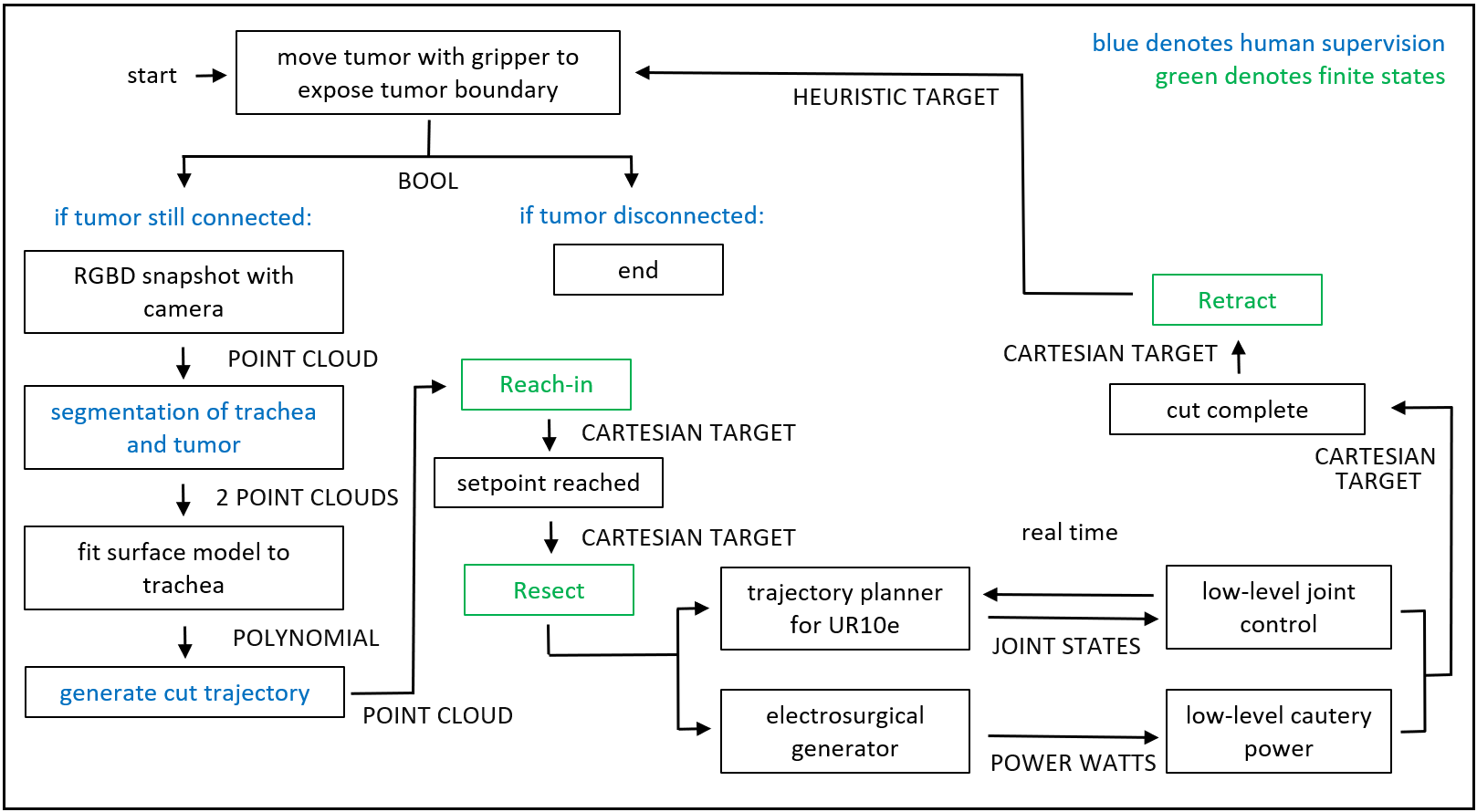}
    \caption{Autonomous control workflow with inputs and outputs for supervised vision-guided resection. Blue denotes events with human supervision and green denotes finite state subtasks.}
    \label{fig:workflow}
\end{figurehere}

\noindent In the \textit{Reach-in} phase, the electrocautery tool approaches the trachea until it is aligned with the trachea's centroid at a predefined safe initial position near the tissue. During the \textit{Resect} phase, the electrocautery tool executes incremental trajectories along the surface of the trachea. 

After each consecutive cut (\textit{Resect} phase), the cautery tool proceeds to a safe home position near the tissue (\textit{Retract} phase), and the tumor is then tensioned by the gripper to expose the tumor boundary. Then, the tissue is re-imaged and re-segmented, and a new surface model of the trachea is generated. A new cut trajectory is generated and displayed to the human supervisor along with updated RMSE values which compare the new surface and cut to the previous surface and cut. Upon supervisor approval, the tool proceeds to \textit{Reach-In} and \textit{Resect}. This pause for supervision enables real-time verification of each cut, providing an additional layer of safety. This workflow is implemented in C++ utilizing ROS 2 for real-time communication, the Point Cloud Library (PCL) for point cloud processing and Eigen for matrix calculations. 

\subsection{Tissue Models}

Due to the size of our rigid-link end-effectors, which limits their operation within a confined trachea, we introduce an ex-vivo \gls{cao} tissue model for an open procedure, comprised of chicken tissue and a `half-pipe' of porcine trachea (Fig. \ref{fig:intro}b, \ref{fig:intro}c). The usage of real animal tissue provides realistic effects from electrocautery and tissue deformation, as well as accurate color and texture information for segmentation, making our findings readily applicable to in-vivo scenarios.

\subsection{Segmentation}
The segmentation pipeline extracts the trachea point cloud from RGBD snapshots to generate the cut trajectory. As shown in Fig.~\ref{fig:sam}, the pipeline combines a custom object detection model based on Faster R-CNN~\cite{ren2015faster}, Meta's Segment Anything model (SAM 1)~\cite{kirillov2023segment}, and 2D mask projection to generate labeled point clouds.

\subsubsection{Object Detection Model}
Based on the Faster R-CNN architecture~\cite{ren2015faster}, the custom object detection neural network produces labeled bounding boxes and their classification scores (CLS Score), for the trachea and tumor. The implementation leverages a pre-trained Faster R-CNN model with ResNet50 FPN backbone~\cite{he2015deep} (trained on the COCO dataset), modified to locate the trachea and tumor. Transfer learning enables a model fine-tuning by replacing the final prediction layer with a custom layer to classify three classes (background, 

\vspace{1em}
\begin{figurehere}
    \centering
    \includegraphics[width=\columnwidth]{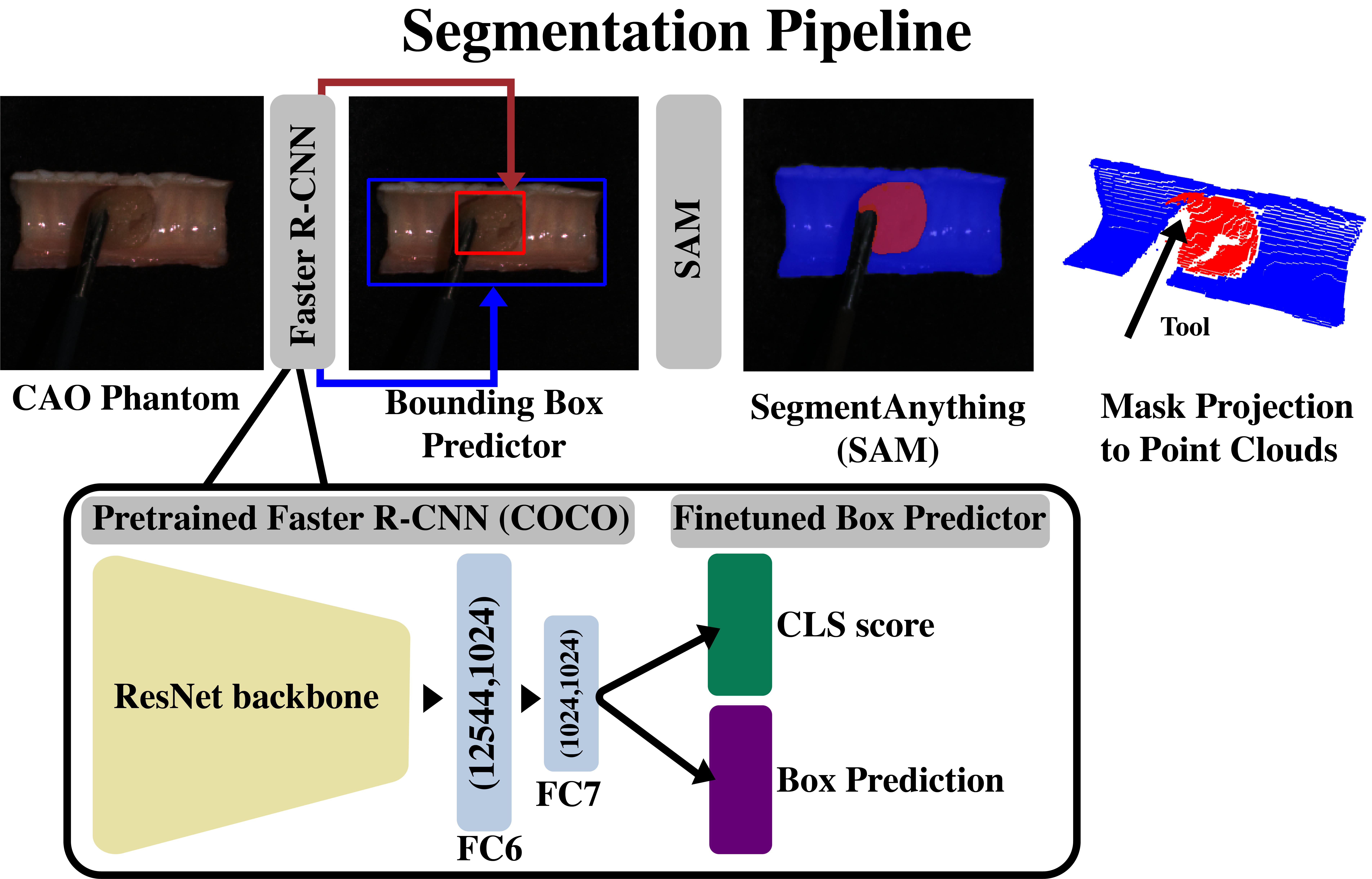}
    \caption{Before each cut, the segmentation pipeline produces labeled point clouds of the trachea and the tumor. }
    \label{fig:sam}
\end{figurehere}

\noindent trachea, and tumor).
The model was trained using stochastic gradient descent (SGD) with momentum (0.9) and weight decay (0.0005) at a learning rate of 0.005 for 10 epochs. The training dataset was manually labeled by a human operator who drew the bounding boxes. The dataset comprised 137 labeled bounding boxes for tissue and trachea, processed in mini-batches of 4 images. Data augmentation was limited to tensor conversion to preserve spatial relationships between the trachea and the tumor. Classification was performed using a softmax function, which converts raw output scores into class scores through exponentiation and normalization. During inference, if the classification score is below $70\%$ we ask the human to provide the bounding box.

\subsubsection{Segmentation and Depth Projection}
Our approach combines Faster R-CNN's automatic localization capabilities with SAM's prompt based foundation model architecture to create an end-to-end segmentation pipeline. Faster R-CNN first predicts the location of tumor and trachea regions, which then serve as learned prompts for SAM to generate precise segmentation masks, eliminating the need for manual bounding box annotations from human supervisors. The generated binary masks from SAM are applied to the depth image, effectively setting all non-masked pixels to zero depth. This operation isolates the region of interest in the depth map. Given a masked depth image and the camera intrinsics, the pinhole camera model projects the pixels to point clouds: 
\begin{align*}
X &= \frac{(u - c_x) \cdot Z}{f_x} \\
Y &= \frac{(v - c_y) \cdot Z}{f_y} \\
Z &= \text{depth\_value}.
\end{align*}
Here, the $X$ and $Y$ coordinates are calculated by subtracting the pixel values $u,v$ from the principal points $c_x, x_y$, which are then divided by the x focal lengths $f_x, f_y$. After projecting the two binary masks to point clouds, the final trachea point cloud is refined by subtracting the tumor point cloud, producing clean, labeled point clouds suitable for surface model fitting. 

\subsection{Trachea Surface Modeling}
Following segmentation, the trachea point cloud has a hole where the tumor is located due to occlusion by the pseudo tumor in the imaging process. We use a surface modeling process to ``fill in" this hole, providing a continuous reference surface for trajectory planning.

To best fit the trachea, various surface fitting models were applied to point clouds taken from four discrete trachea models (Fig.~\ref{fig:polynomials}). The objective  was to determine the optimal fitting models by minimizing two criteria: computation time and root mean squared error (RMSE). A Pareto front analysis was performed for each trachea model to identify the best models based on the trade-off between these two objectives. To model the trachea surface accurately, we applied polynomial surface fitting models to point cloud data obtained from four different ex-vivo trachea models (captured in .pcd format). A series of polynomial surface fitting models, ranging from degree 1x1 (poly11) to 10x10 (poly1010), were applied to approximate the shape of the trachea. The fitting process involved solving a least-squares regression problem, where polynomial coefficients were determined from the trachea point cloud. Once the polynomial surfaces were fitted, we computed RMSE as a measure of how well each fitted surface matched the original trachea point cloud. This provided a quantitative assessment of how accurately each polynomial model

\vspace{1em}
\begin{figurehere}
    \centering
    \includegraphics[width=0.95\columnwidth]{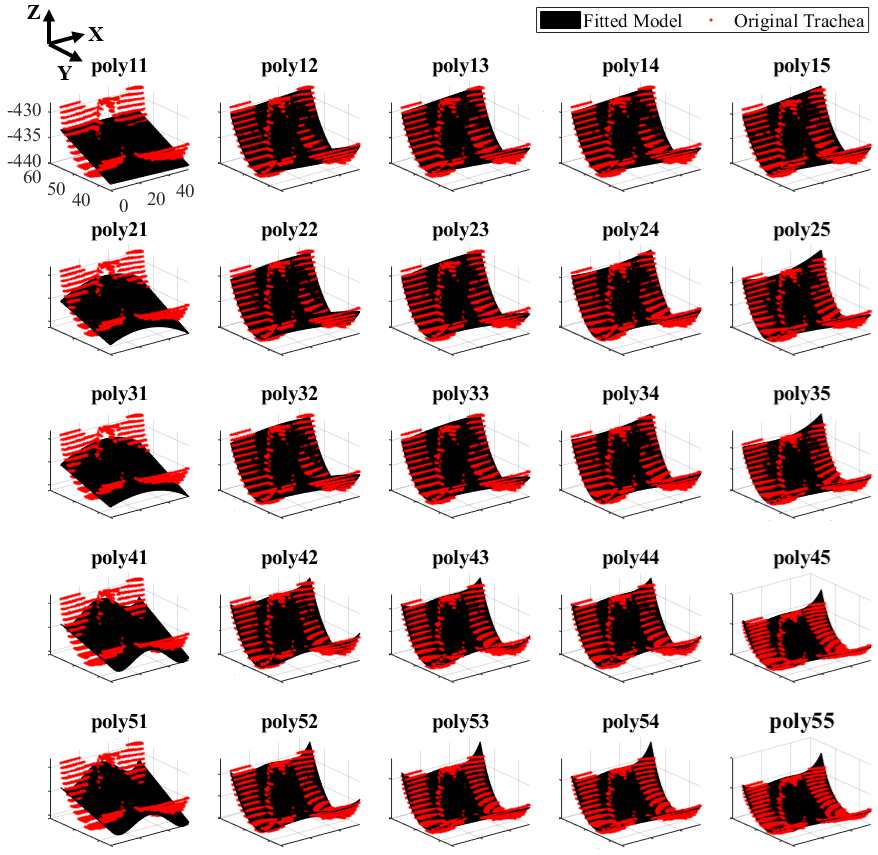}
    \caption{Trachea Model - Polynomial Fits ($poly11$ to $poly55$) for one of the samples (Units are in millimeters).}
    \label{fig:polynomials}
\end{figurehere}

\noindent approximated the real trachea geometry. After computing RMSE separately for all four trachea models, we took the mean RMSE across the four models for each polynomial. In Fig.~\ref{fig:polynomials}, the subplot grid layout is organized by polynomial orders, where the columns represent the increasing polynomial degree in the Y direction and the rows represent the increasing polynomial degree in the X direction.

In addition to the error, the computation time required to fit each surface to the trachea data was recorded. This served to evaluate the computational efficiency of each model. After obtaining the RMSE and time values for all models, a Pareto front analysis was conducted, the results of which are shown in Fig.~\ref{fig:paretoFront}. In Fig.~\ref{fig:paretoFront}, each point represents a polynomial model plotted based on its RMSE and computation time. The models closer to the Pareto front (red dots) offer an optimal balance between accuracy and efficiency.

Even the smallest differences in RMSE are of critical importance, particularly in the context of autonomous surgery. When examining the Pareto optimal polynomials with the mean RMSE values below 1 mm, it is observed that the mean RMSE does not significantly decrease beyond \textit{poly55}, stabilizing around 0.6 mm. Among the tested models, the 5th-order polynomial (\textit{poly55}) achieved the lowest computation time and was therefore chosen as the optimal model. Using \textit{poly55}, the polynomial surface model is expressed as:
\begin{equation}
    P(x, y) = \sum_{i=0}^{5} \sum_{j=0}^{5} a_{ij} x^i y^j \quad \text{where} \quad i + j \leq 5.
    \label{eq:5thOrder} \nonumber
\end{equation}
Here, $a_{ij}$ are the polynomial coefficients to be determined. This polynomial expression defines the surface by capturing the relationship between the $x$ and $y$ coordinates and the resulting $z$ values (surface height). To obtain the coefficients, the known $x$ and $y$ values are substituted into $P(x, y)$, and the resulting system of equations is solved by equating the polynomial output to the corresponding $z$ values from the point cloud. This fitting process generates a polynomial surface that closely models the trachea, where $i$ and $j$ are non-negative integers up to 10, with each term $x^i y^j$ representing a combination of polynomial degrees in $x$ and $y$.\looseness-1

\vspace{1em}

\begin{figurehere}
    \centering
    \includegraphics[width=0.95\columnwidth]{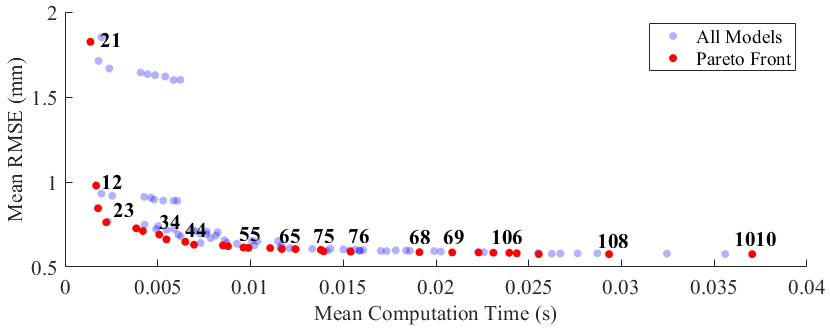}
    \caption{Pareto front analysis of polynomial surface fitting models for trachea modeling. Each point represents a polynomial model, plotted according to its computation time and root mean squared error (RMSE). Models closer to the Pareto front (red dots) achieve an optimal trade-off between computational efficiency and accuracy.}
    \label{fig:paretoFront}
\end{figurehere}


\subsection{Electrocautery Pitch Angle and Trajectories} 

To determine what pitch angle to set the electrocautery tool during the autonomous resection, four handheld surgical procedure demonstrations (with manual cautery and a laparoscopic gripper) were analyzed. In each demonstration, a point cloud snapshot of the surgical scene was taken during each of the first four cuts. When these point clouds were viewed in the x-z plane, the angle of the electrocautery tool could be determined by fitting a linear model to points along the tool, as shown in Fig. \ref{fig:cautery-pitch}. From these demonstrations, a mean electrocautery pitch of 28.3\(^\circ\) $\pm$ 4.6\(^\circ\) was calculated (Table~\ref{cutpitch}). This informed the programmed pitch angle of the electrocautery tool during the autonomous resection.

\vspace{1em}
\begin{figurehere}
    \centering
    \includegraphics[width=0.95\columnwidth]{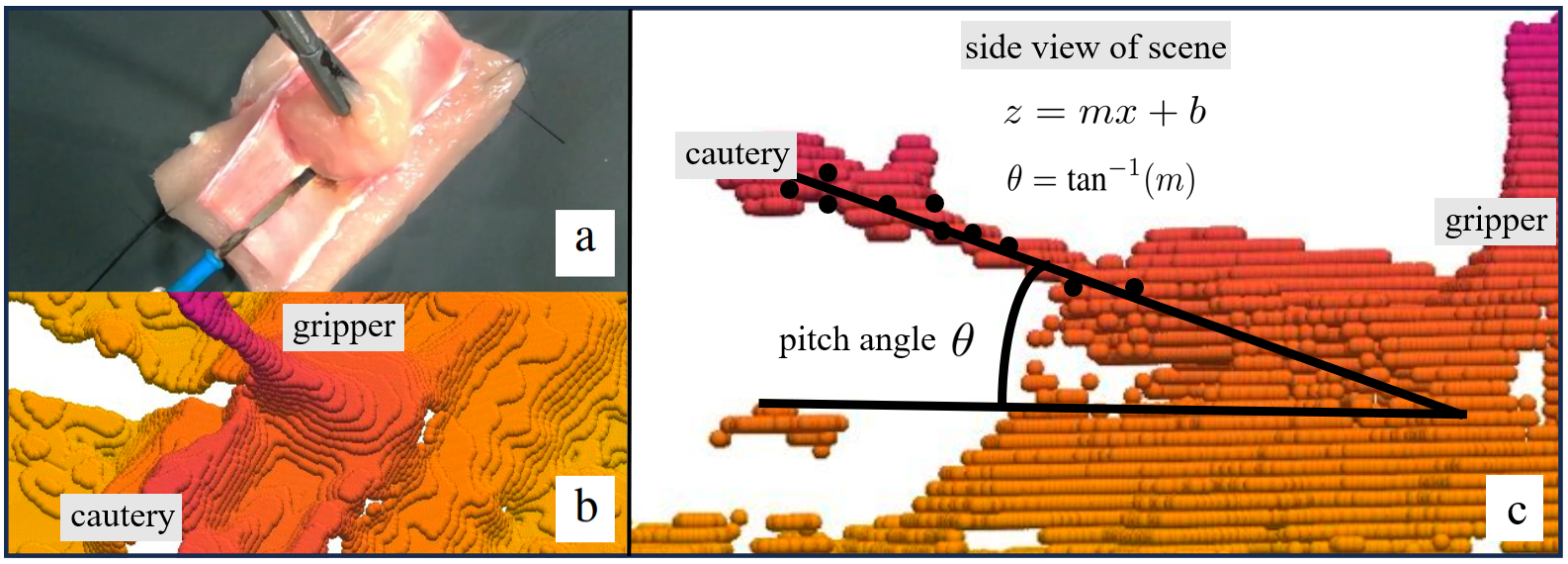}
    \caption{Handheld demonstration of CAO resection (a), resulting point cloud snapshot during the first cut (b), and the X-Z plane of the point cloud and linear model (c).}
    \label{fig:cautery-pitch}
\end{figurehere}

\begin{tablehere}
\centering
\begin{tabular}{|c|c|c|c|c|}  \hline
     & Cut 1 & Cut 2 & Cut 3 & Cut 4  \\ \hline
  Model 1 & 21.7\(^\circ\)     & 20.4\(^\circ\)     & 24.7\(^\circ\)     & 29.1\(^\circ\)     \\ \hline
  Model 2 & 20.8\(^\circ\)     & 29.4\(^\circ\)     & 34.1\(^\circ\)     & 36.5\(^\circ\)     \\ \hline
  Model 3 & 28.7\(^\circ\)     & 27.9\(^\circ\)     & 28.2\(^\circ\)     & 29.2\(^\circ\)     \\ \hline
  Model 4 & 32.0\(^\circ\)     & 27.0\(^\circ\)     & 31.6\(^\circ\)     & 31.4\(^\circ\)     \\ \hline
             \multicolumn{5}{|c|}{Mean $\pm$ STD = 28.3\(^\circ\) $\pm$ 4.6\(^\circ\)} \\ \hline
\end{tabular}
\caption{Electrocautery pitch for the first four cuts, within four distinct handheld demonstrations.}
\label{cutpitch}
\end{tablehere}

For each trajectory generation, the generated surface fit of the trachea is utilized. The trajectory is comprised of six total cut paths, which are designed to be evenly spaced over the size of the tumor in the X-Y plane. The number of cut paths is currently fixed to the arbitrary value of 6 cuts; in the future this could be modified such that the number of cuts varies based on tumor size. For each cut path, the Z values are determined using the polynomial surface model of the trachea, with an added offset of 1 mm as not to puncture the trachea surface, thus defining the trajectory path 1 mm above the real trachea surface. Each cut path is traveled left-to-right by the electrocautery tool, and then retraced right-to-left, after which the electrocautery tool returns to a pulled-back home position. Fig.~\ref{fig:traj} illustrates the fitted surface and cut paths as simulated in RViz. In cycle \textit{i}, the program receives the tumor cloud \textit{i} and trachea surface \textit{i}, and plans cut paths \textit{i} and \textit{i+1}. The robot then executes cut path \textit{i}, and the tumor is then retracted backward. Then, in cycle \textit{i+1}, the program receives new data for the tumor cloud \textit{i+1} and trachea surface \textit{i+1}. It then plans cut paths \textit{i+1} and \textit{i+2}. Before executing the cut path \textit{i+1}, the program first compares the RMSE errors between the surface \textit{i+1} created in the first cycle \textit{i}, to the surface \textit{i+1} created in the second cycle \textit{i+1}. The RMSE check serves as a self-supervision mechanism to detect significant discrepancies in trajectory predictions. If RMSE is low, the next cut follows the predicted path without human intervention. If RMSE is high, this suggests segmentation errors, tissue deformation or unexpected tool interactions, prompting a human supervisor to validate the cut before execution. Sometimes, errors in surface modeling can occur due to inaccurate segmentation. In such cases, human supervision assesses the RMSE to determine whether the trajectory for cut \textit{i+1} is suitable. This process continues for all cuts.

\vspace{1em}
\begin{figurehere}
    \centering
    \includegraphics[width=\columnwidth]{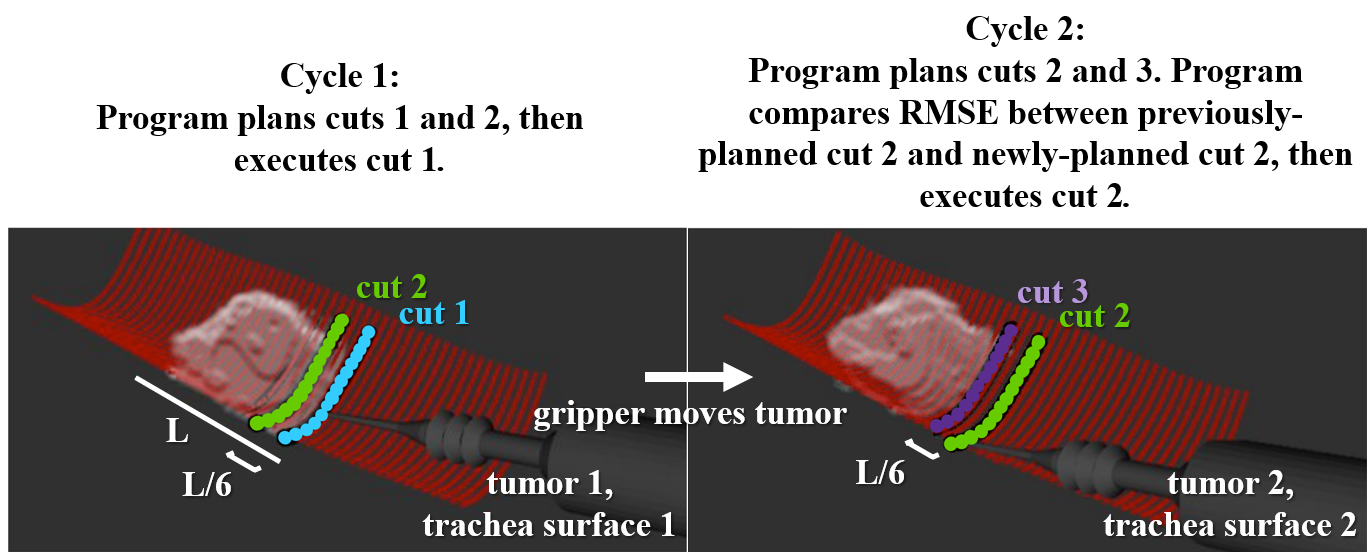}
    \caption{Fitted surface model and successive cutting trajectories in simulation (RViz). Between cycles, the tumor is retracted backward.}
    \label{fig:traj}
\end{figurehere}   
\section{Experiments}

We tested our autonomous resection workflow by conducting consecutive procedures on five distinct tissue models. First, five \gls{cao} tissue models were fabricated to mimic a range of clinical cases such that they are somewhat varied in shape and size. Each tissue model was created by first sculpting a piece of chicken tissue (Giant Food, Carlisle, PA) such that the base measures approximately 50\,mm by 75\,mm, and the attached tumor protrusion is approximately 20\,mm in diameter. The chicken tissues, which were stored in a freezer prior to the model fabrication, tended to release moisture over time, making them challenging to grasp securely with the standard gripper. To address this, fresh chicken tissues were used, and allowed to dry slightly before beginning the procedure. Ex-vivo porcine tracheas (Animal Technologies, Tyler, TX) were cut into `half-pipes' and then into lengths of approximately 75\,mm. A hole was cut in the center of each trachea. Each trachea was lowered down onto the chicken model, and secured into place with Loctite 415 (McMaster-Carr, Elmhurst, IL) and Insta-Set accelerator (Bob Smith Industries, Paso Robles, CA). The wide base made of chicken tissue provided ample contact area between each model and the electrosurgical grounding plate, enabling steady monopolar electrocautery. After all five models were fabricated, preliminary CT scans were taken of each model. The ground-truth segmentations of the tumors were then obtained from the CT (using 3D Slicer~\cite{Fedorov2012}), as shown in Fig.~\ref{fig:resection-results}. 

To begin each procedure, a tissue model was placed onto the electrosurgical grounding plate and secured in place with four stay sutures. Two pieces of black felt cloth were placed under the trachea to obfuscate the background from the camera view for ease of segmentation. The Zivid camera first captured a depth image, RGB image, camera intrinsics, and 3D point cloud of the surgical scene. Using this data, a custom Faster R-CNN model generated bounding boxes and SAM-based segmentation masks which were projected onto the depth data. A fifth-degree polynomial (poly55) was fitted to the segmented trachea to generate smooth cut trajectories. The robot then began the Reach-in stage, aligning with the trachea’s centroid, followed by the Resect stage, executing cuts at 2 mm/s with 24 W electrosurgical power. A Smoke Shark (Bovie, Clearwater, FL) evacuated cautery smoke. After each cut, the robot entered the Retract stage to return to a safe position. A manually positioned laparoscopic gripper then tensioned the tumor to expose the cut boundary. The cut was then shifted by L/6 mm for subsequent resections, where L is the tumor length along the trachea. Process images taken during the experiment are shown in Fig. 8. The procedure was stopped when the tumor was detached, and a postoperative CT scan evaluated resection effectiveness.

\vspace{1em}
\begin{figurehere}
    \centering
    \includegraphics[width=\columnwidth]{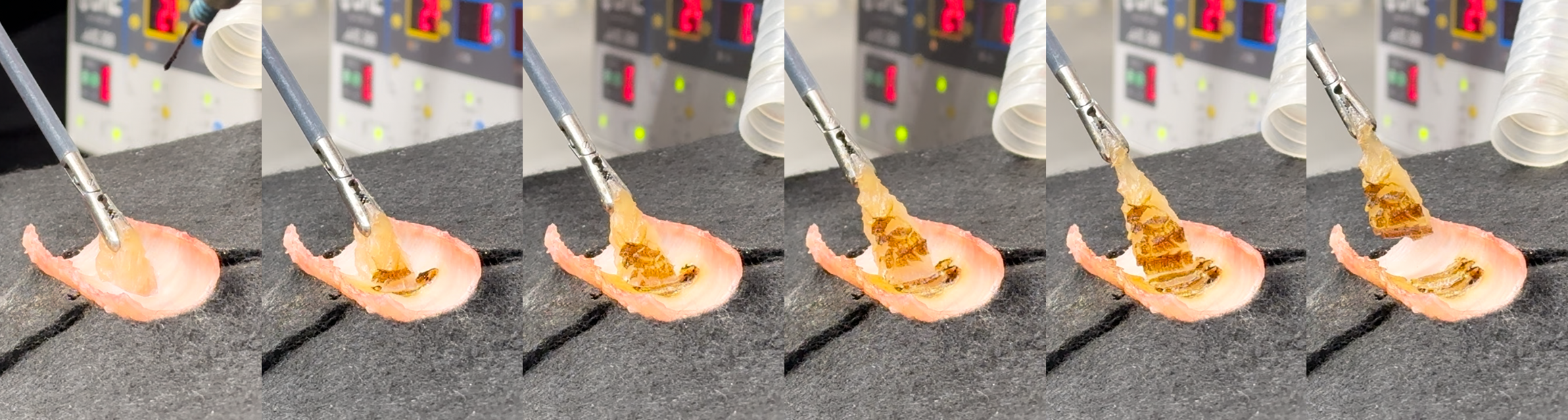}
    \caption{Snapshots taken from an experimental video showing the full resection process for Model 2.}
    \label{fig:resection-process}
\end{figurehere}
\vspace{1em}

\subsection{Results}

\begin{figure*}[ht!]
    \centering
    \includegraphics[width = 0.7\textwidth]{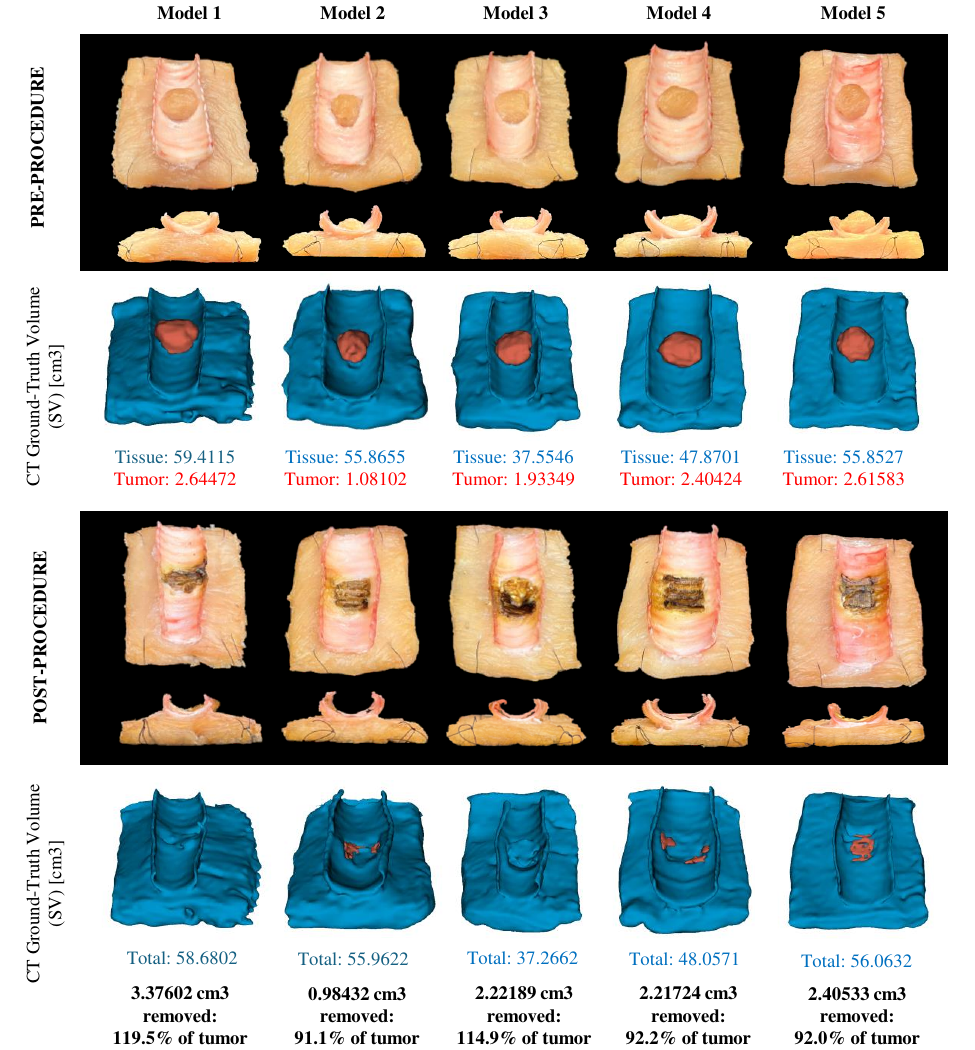}
    \caption{Results of resection experiments on 5 discrete tissue models. Pre-procedural top and side views are shown, as well as the pre-procedural ground-truth CT and segmentation volumes. Post-procedural top and side views are shown, as well as the post-procedural ground-truth CT.}
    \label{fig:resection-results}
\end{figure*}

All five \gls{cao} models and their corresponding post-procedural outcomes are illustrated in Fig.~\ref{fig:resection-results}. By qualitatively comparing the side views provided, it is evident that the tumor was successfully removed from the trachea in all five consecutive trials. Models 1 and 3 were over-resections (119.5\% and 114.9\% of tumor removed, respectively), while Models 2, 4, and 5 were under-resections (91.1\%, 92.2\%, and 92.0\% of tumor removed, respectively). In this context, it is not necessary to resect a margin around the tumor, since CAO removal is a palliative and not curative procedure. Mudambi et al. defines successful CAO removal as a reopening of the airway lumen to over 50\% of the nominal diameter \cite{Mudambi2017}. By this definition, all five of our procedures can be deemed successful, marking feasibility demonstrations of autonomous vision-based \gls{cao} resection in 5 out of 5 trials.

Additionally, we analyzed the surfaces of the models after each procedure. After each resection was complete, a snapshot was taken of the tissue model, and the regular point cloud segmentation method was conducted. A surface model was fit to the trachea point cloud and then raised in Z by 1\,mm to create the 'goal' surface (which, according to the programmed trajectories, should align closely with the charred top of the tumor). A point cloud of the charred area was also obtained. For each point in the charred area, the Z-value was compared to the corresponding Z-value of the 'goal' surface. In this way, RMSE could be calculated between the programmed cut surface and the real cut surface seen post-procedure. The tissue surfaces for the actual tissue (charred tissue post-cut) and fitted tissue (based on trachea surface) for all five models are shown in Fig. \ref{fig:surface-fit}. The RMSE for each model is also provided, which range from a minimum of 1.23\,mm to a maximum of 2.65\,mm.

\vspace{1em}
\begin{figurehere}
    \centering
    \includegraphics[width=\columnwidth]{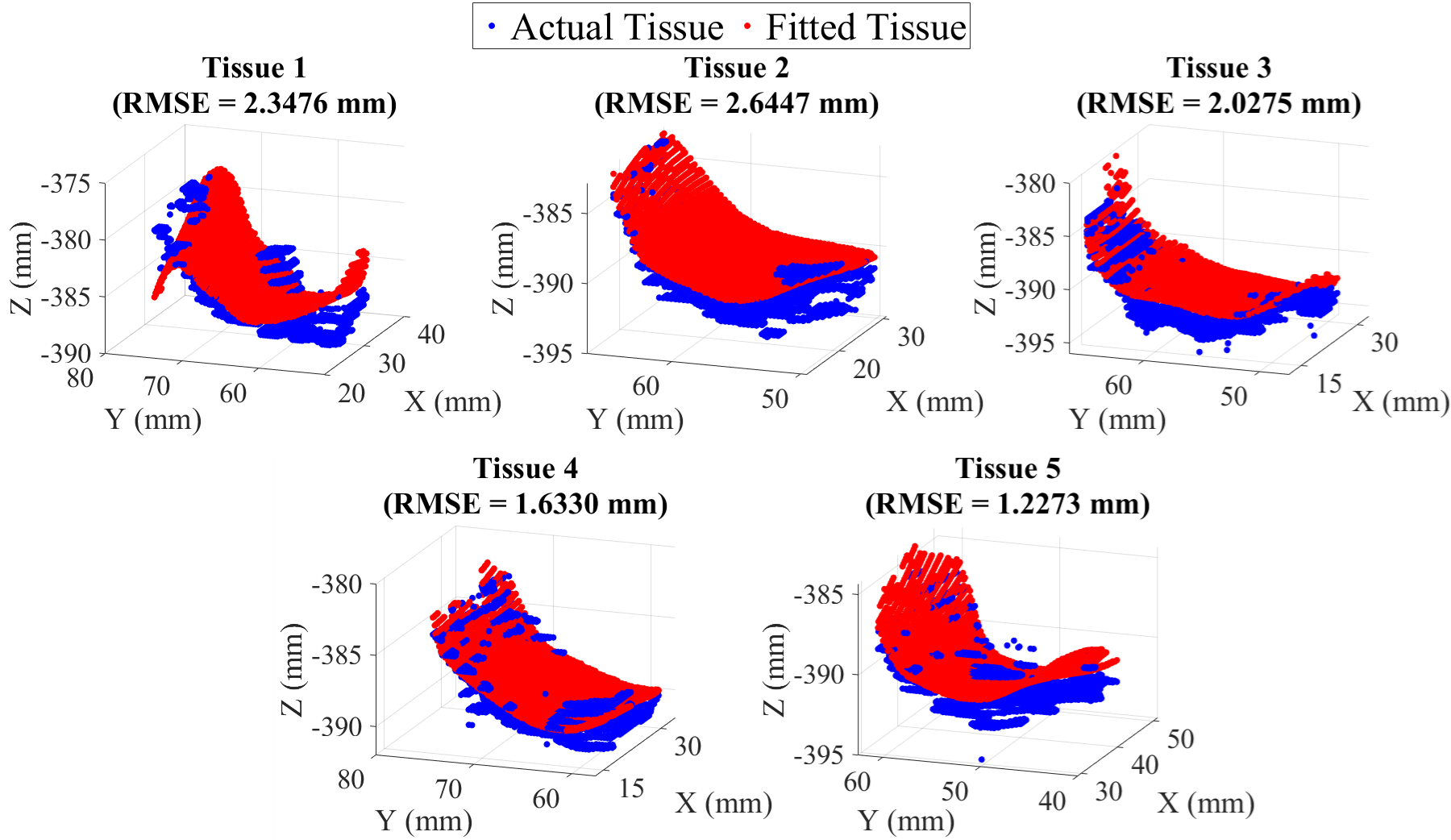}
    \caption{Plotted surfaces for the actual tissue (charred tissue post-cut) and fitted tissue (our goal from trachea surface fit) for all five models. RMSE is calculated for each model.}
    \label{fig:surface-fit}
\end{figurehere}

We also analyzed the effectiveness of our segmentation pipeline through the five procedures. We found that the ability to extract accurate point clouds relied more on the ability to predict bounding boxes than on the resulting segmentation of the objects (which leverages the predicted bounding boxes). Our custom Faster R-CNN model outperformed simple CNNs in terms of both intersection over union (IoU) scores and the downstream task of producing an accurate trachea point cloud, as shown in Fig.~\ref{fig:IoU}. The faster R-CNN had an IoU of $(0.762 \pm 0.283)$ for the trachea and $(0.592 \pm 0.36)$ for the tumor. The simple CNN had IOU scores of $(0.49 \pm 0.2567)$ for the trachea and $(0.150 \pm 0.181)$ for the tumor (higher is better for IoU). 

\begin{figurehere}
    \centering
    \includegraphics[width=\columnwidth]{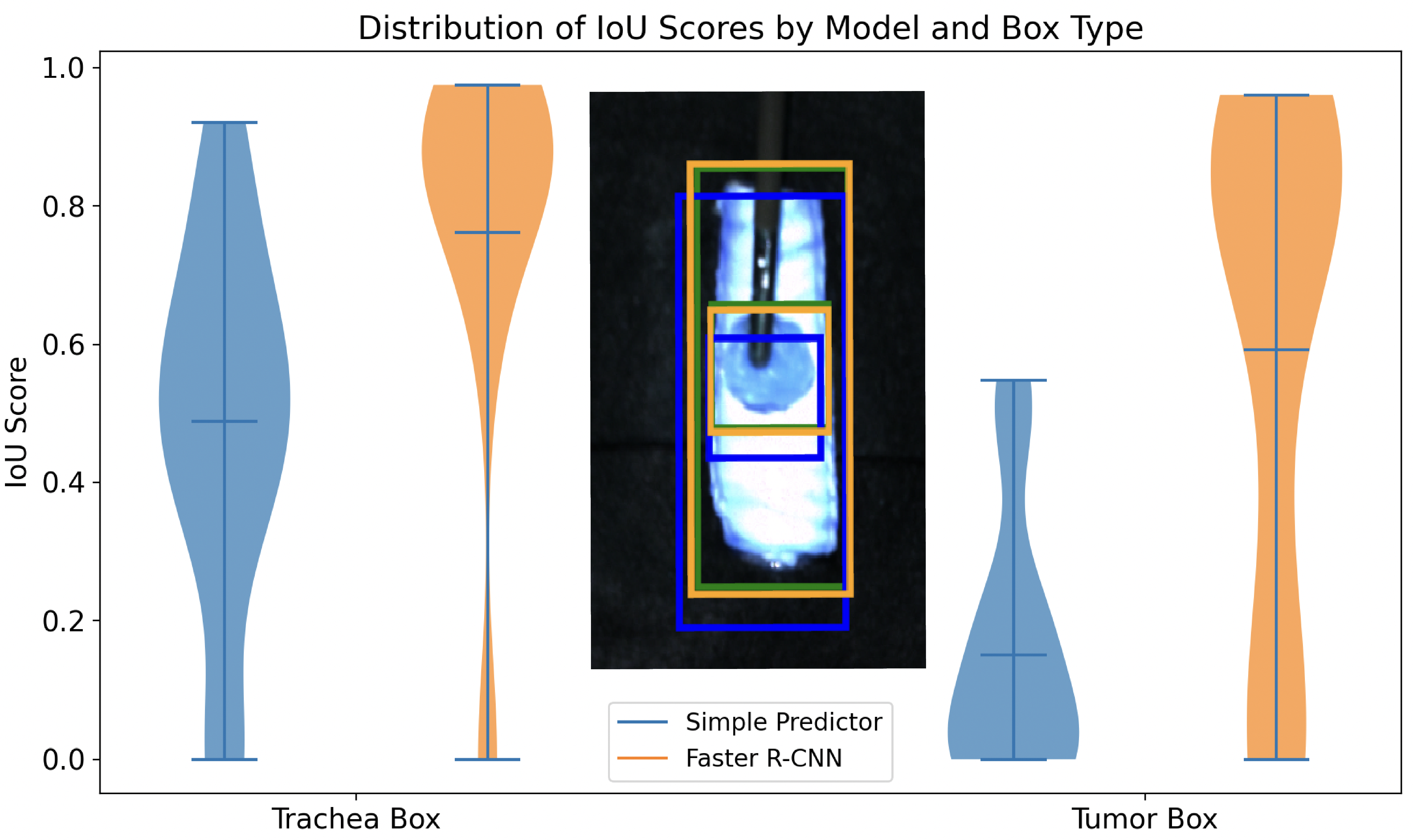}
    \caption{Violin plots showing the Intersection over Union scores for our custom Faster R-CNN model versus a simple CNN over the five procedures for the trachea (left) and tumor (right), with a representative example shown in the center. The orange bounding box is the Faster R-CNN, blue is the simple CNN, and green is the ground truth. }
    \label{fig:IoU}
\end{figurehere}

\vspace{1em}

\section{Discussion}

As mentioned in the Results, the clinical benchmark for successful resection of CAO is a reopening of the airway lumen to over 50\% of its nominal diameter. Our resections were successful by this definition. Moreover, our results indicate a potentially superior clinical outcome of lumen reopening by over 90\%. From a technical standpoint, we had set an initial target to leave a 1\,mm cutting offset above the trachea. This targeted offset was intended to remove approximately 90\% of the tumor, leaving the remaining 10\% to avoid damaging the trachea. However, due to calibration limitations and minor bending of the electrocautery tool during the procedure, the system occasionally deviated from this target. It is also a possibility that those deviations were caused in part by the relative spatial inaccuracies of the UR robots (0.1\.mm repeatability for the UR10e). In some trials, the tumor was under-resected as intended, but in others, the tumor was over-resected because the electrocautery tool penetrated into the trachea. In future work, it will be key to conduct a more exact calibration, use a more accurate robotic system, or possibly switch to a more rigid electrocautery tool, to permit higher accuracy in resection.

The tool-tissue interactions between the chicken and gripper also posed challenges during the experiments. When the chicken was too wet, it easily slipped out of the gripper. The chicken also began to tear after being gripped for an extended period of time. In future work, we hope to employ an automated gripping strategy rather than the passive heuristic gripping used in this study. An autonomous gripper may be able to detect tissue slipping and prompt a re-gripping motion, and it could perhaps also modulate the applied force. As we look toward employing our workflow on minimally-invasive robots in the future, a possible solution for gripping could be to instead push and prod the tumor away from the cut, rather than biting it with a toothed gripper. This would be advantageous since it would likely cause less tissue damage, would require less applied force, and would allow the minimally-invasive gripper to approach from the distal (mouth) end of the trachea.

Considering the minimally-invasive translation, it is relevant to reconsider the position of the camera in our configuration. In this study, we placed the camera above the tissue, since the robotic components were too large to place the camera on the wrist of the electrocautery tool. However, in a minimally invasive case, the electrocautery tool, gripper, and camera would all approach together from the distal (mouth) end. While a camera approaching from the distal end would have a much different perspective than the camera in this study, we hypothesize that a distal camera would actually be advantageous for our resection task. A distal camera would be able to see directly into the cuts being made, and could perhaps provide a more accurate representation of the boundary between the tumor and the trachea. If translated to a minimally-invasive robot, the camera would likely take the form of some monocular endoscope, with depth information inferred from the endoscopic footage using techniques such as SLAM and SFM \cite{Masoumian2022, Liu2022}.

There were also some challenges present in our segmentation network. The intermediate stages of tumor removal were not significantly represented in the training data, leading to certain faults during segmentation. For example, tissue charring from electrocautery was occasionally detected as part of the tumor, necessitating the human supervisor to replace the predicted bounding boxes with manually drawn ones. During the procedures presented in this study, the human chose to intervene to draw manual boxes for 5 out of the 27 total segmentation steps. Also, the segmentation process occasionally included a part of the gripper in the trachea mask, leading to outliers in the trachea point cloud. This led to errors in the surface model and consequently in the generated trajectories, particularly along the Z-axis. This misassignment resulted in inaccurate depth estimations, affecting the precision of the planned cuts and increasing the possibility of unintended interaction with the trachea. In future work, it will be necessary to include a proper distribution of mid-procedure charred data in the training data for segmentation.

Most importantly, our current resection architecture is designed as an open-loop system. As future work, we aim to develop a closed-loop framework that incorporates a feedback mechanism to adapt to tissue deformations in real time. Such a system would enable the resection process to dynamically respond to changes in the surgical environment. 

To translate this research to minimally-invasive manipulators which can fit within the trachea, our workflow would remain viable, with the snapshot and segmentation processes replaced by endoscopic monocular SLAM \cite{Masoumian2022, Liu2022}. To incorporate autonomous functions into surgical workflows, systems like the da Vinci robot could suggest cutting plans, similar to a "park assist" feature in cars. The robot would analyze the surgical site and propose actions, but the surgeon could easily intervene at any point. To ensure safety and effectiveness, the system would allow real-time overrides and provide transparent reasoning for its suggestions.

\vspace{12pt}
\nonumsection{Acknowledgments}

This material is supported in part by the Advanced Research Projects Agency for Health (ARPA-H) under grant number D24AC00415, and by the NSF Foundational Research in Robotics (FRR) Faculty Early Career Development Program (CAREER) under grant number 2144348. Any opinions, findings, and conclusions or recommendations expressed in this material are those of the authors and do not necessarily reflect the views of ARPA-H and NSF.

\bibliographystyle{ws-jmrr}
\bibliography{CAO-References.bib}

\begin{thebibliography}{10}

\bibitem{Chen2011}
E.~Chan, Malignant airway obstruction: treating central airway obstruction in the oncologic setting, {\em UWOMJ} {\bf 80}  (2011)  7--9.

\bibitem{Chen1998}
K.~Chen, J.~Varon and O.~C. Wenker, Pii s0736-4679(97)00245-x selected topics: Critical care malignant airway obstruction: Recognition and management  (1998).

\bibitem{Ernst2004}
A.~Ernst, D.~Feller-Kopman, H.~D. Becker and A.~C. Mehta, Central airway obstruction (6 2004).

\bibitem{Morris2002}
C.~D. Morris, J.~M. Budde, K.~D. Godette, T.~L. Kerwin, J.~I. Miller and J.~B. Whitehead, Palliative management of malignant airway obstruction  (2002).

\bibitem{Mathisen1989}
D.~J. Mathisen and H.~C. Grillo, Endoscopic relief of malignant airway obstruction, {\em Annals of Thoracic Surgery} {\bf 48}  (1989)  469--475.

\bibitem{Vishwanath2013}
G.~Vishwanath, K.~Madan, A.~Bal, A.~N. Aggarwal, D.~Gupta and R.~Agarwal, Rigid bronchoscopy and mechanical debulking in the management of central airway tumors an indian experience  (2013).

\bibitem{Gafford2020}
J.~B. Gafford, S.~Webster, N.~Dillon, E.~Blum, R.~Hendrick, F.~Maldonado, E.~A. Gillaspie, O.~B. Rickman, S.~D. Herrell and R.~J. Webster, A concentric tube robot system for rigid bronchoscopy: A feasibility study on central airway obstruction removal, {\em Annals of Biomedical Engineering} {\bf 48} (1 2020)  181--191.

\bibitem{Kehoe2014}
B.~Kehoe, G.~Kahn, J.~Mahler, J.~Kim, A.~Lee, A.~Lee, K.~Nakagawa, S.~Patil, W.~D. Boyd, P.~Abbeel and K.~Goldberg, Autonomous multilateral debridement with the raven surgical robot.

\bibitem{Hu2018}
D.~Hu, Y.~Gong, E.~J. Seibel, L.~N. Sekhar and B.~Hannaford, Semi-autonomous image-guided brain tumour resection using an integrated robotic system: A bench-top study, {\em International Journal of Medical Robotics and Computer Assisted Surgery} {\bf 14} (2 2018).

\bibitem{McKinley2016}
S.~McKinley, A.~Garg, S.~Sen, D.~V. Gealy, J.~P. McKinley, Y.~Jen, M.~Guo, D.~Boyd and K.~Goldberg, An interchangeable surgical instrument system with application to supervised automation of multilateral tumor resection, {\em IEEE International Conference on Automation Science and Engineering\/},   {\bf 2016-November}, (IEEE Computer Society, 11 2016), pp. 821--826.

\bibitem{Opfermann2017}
J.~D. Opfermann, S.~Leonard, R.~S. Decker, N.~A. Uebele, C.~E. Bayne, A.~S. Joshi and A.~Krieger, Semi-autonomous electrosurgery for tumor resection using a multi-degree of freedom electrosurgical tool and visual servoing, {\em IEEE International Conference on Intelligent Robots and Systems\/},   {\bf 2017-September}, (Institute of Electrical and Electronics Engineers Inc., 12 2017), pp. 3653--3660.

\bibitem{Ge2019}
J.~Ge, H.~Saeidi, J.~D. Opfermann, A.~S. Joshi and A.~Krieger, Landmark-guided deformable image registration for supervised autonomous robotic tumor resection, {\em Lecture Notes in Computer Science (including subseries Lecture Notes in Artificial Intelligence and Lecture Notes in Bioinformatics)\/},   {\bf 11764 LNCS}, (Springer Science and Business Media Deutschland GmbH, 2019), pp. 320--328.

\bibitem{Ge2021}
J.~Ge, H.~Saeidi, M.~Kam, J.~Opfermann and A.~Krieger, Supervised autonomous electrosurgery for soft tissue resection, {\em BIBE 2021 - 21st IEEE International Conference on BioInformatics and BioEngineering, Proceedings\/},  (Institute of Electrical and Electronics Engineers Inc., 2021).

\bibitem{Ge2024}
J.~Ge, M.~Kam, J.~D. Opfermann, H.~Saeidi, S.~Leonard, L.~J. Mady, M.~J. Schnermann and A.~Krieger, Autonomous system for tumor resection (astr) - dual-arm robotic midline partial glossectomy, {\em IEEE Robotics and Automation Letters} {\bf 9} (2 2024)  1166--1173.

\bibitem{Nagy2018}
D.~Ákos Nagy, T.~D. Nagy, R.~Elek, I.~J. Rudas and T.~Haidegger, Ontology-based surgical subtask automation, automating blunt dissection, {\em Journal of Medical Robotics Research} {\bf 3} (9 2018).

\bibitem{FSMs}
R.~Balogh and D.~Obdr{\v{z}}{\'a}lek, Using finite state machines in introductory robotics, {\em Robotics in Education\/},  eds. W.~Lepuschitz, M.~Merdan, G.~Koppensteiner, R.~Balogh and D.~Obdr{\v{z}}{\'a}lek (Springer International Publishing, Cham, 2019), pp. 85--91.

\bibitem{ren2015faster}
S.~Ren, K.~He, R.~Girshick and J.~Sun, Faster r-cnn: Towards real-time object detection with region proposal networks  (2015).

\bibitem{kirillov2023segment}
A.~Kirillov, E.~Mintun, N.~Ravi, H.~Mao, C.~Rolland, L.~Gustafson, T.~Xiao, S.~Whitehead, A.~C. Berg, W.-Y. Lo, P.~Dollár and R.~Girshick, Segment anything  (2023).

\bibitem{he2015deep}
K.~He, X.~Zhang, S.~Ren and J.~Sun, Deep residual learning for image recognition, {\em arXiv preprint arXiv:1512.03385}   (2015).

\bibitem{Fedorov2012}
A.~Fedorov, R.~Beichel, J.~Kalpathy-Cramer, J.~Finet, J.~C. Fillion-Robin, S.~Pujol, C.~Bauer, D.~Jennings, F.~Fennessy, M.~Sonka, J.~Buatti, S.~Aylward, J.~V. Miller, S.~Pieper and R.~Kikinis, 3d slicer as an image computing platform for the quantitative imaging network, {\em Magnetic Resonance Imaging} {\bf 30} (11 2012)  1323--1341.

\bibitem{Mudambi2017}
L.~Mudambi, R.~Miller and G.~A. Eapen, Malignant central airway obstruction (9 2017).

\bibitem{Masoumian2022}
A.~Masoumian, H.~A. Rashwan, J.~Cristiano, M.~S. Asif and D.~Puig, Monocular depth estimation using deep learning: A review (7 2022).

\bibitem{Liu2022}
X.~Liu, Z.~Li, M.~Ishii, G.~D. Hager, R.~H. Taylor and M.~Unberath, Sage: Slam with appearance and geometry prior for endoscopy, {\em Proceedings - IEEE International Conference on Robotics and Automation\/},  (Institute of Electrical and Electronics Engineers Inc., 2022), pp. 5587--5593.

\end{thebibliography}

\end{multicols}
\end{document}